\documentclass[10pt,twocolumn]{article}

\usepackage[T1]{fontenc}
\usepackage[letterpaper,margin=0.72in,columnsep=0.25in]{geometry}
\usepackage{newtxtext}
\usepackage[hyphens]{url}
\usepackage{graphicx}
\usepackage{natbib}
\usepackage{caption}
\usepackage{algorithm}
\usepackage{algorithmic}
\usepackage{amsmath,amssymb}
\usepackage{booktabs}
\usepackage{microtype}
\usepackage[colorlinks=true,allcolors=blue]{hyperref}

\newcommand{\pg}{\textsc{pg}}

\title{Compile, Then Page: Executable SOP Programs and a\\
Capability-Gated Runtime for Procedural LLM Agents}

\author{
Chenglin Yu$^{1}$, Li Yin$^{2}$, Qingxin Fan$^{4}$, Ying Yu$^{3}$,\\
Runyang Ray Zhong$^{2}$, Ming Li$^{1,5}$\\[2pt]
\small $^{1}$Department of Industrial and Systems Engineering, The Hong Kong Polytechnic University\\
\small $^{2}$Department of Data and Systems Engineering, The University of Hong Kong\\
\small $^{3}$College of Economics and Management, Zhejiang Normal University\\
\small $^{4}$Shanghai Juepei Flexible Supply Chain Technology Co., Ltd.\\
\small $^{5}$Research Institute for Generative AI, The Hong Kong Polytechnic University\\
\small Correspondence: ming.li@polyu.edu.hk
}
\date{}

\begin{document}
\maketitle

\begin{abstract}
Enterprise agents must follow long-horizon, conditional, safety-critical standard operating procedures (SOPs). We compile machine-readable SOP constraints into executable pseudo-code and run them with a program-guided (PG) stack machine that pages the active frame while an LLM performs semantic execution. A three-arm SOPBench study across six models separates representation from runtime: compiled text never significantly hurts and gains up to 16.0 points where official prose underperforms. Runtime guidance is capability-gated. Two strong models independently show positive seven-domain PG contrasts (58:19 and 75:31 discordant pairs), whereas weak models are harmed. A full-program cursor ablation---active frame first, complete program retained---recovers much of the strong-model refusal gain; selective visibility adds a smaller improvement. Paired probe and audit measurements track this divide to \emph{spontaneous} state discipline rather than reconstruction ability. On Bank the three primary arms rise 70.4$\rightarrow$86.4$\rightarrow$92.8, with 100\% refusal correctness. Practical guidance: compile first; enable active-frame paging only after a model-level discipline check.

\end{abstract}

\section{Introduction}
\label{sec:intro}

Deploying language agents in customer-facing operations is, to a first approximation, a procedure-following problem. Banks, clinics, and service desks encode their obligations as standard operating procedures (SOPs): long-horizon workflows with conditional branches and safety-critical refusal rules---what to do, what to verify first, when to decline. On SOPBench~\citep{sopbench2025}, the best official configuration passes 82.4\% of Bank tasks on the subset we study, with the best refusal accuracy reaching 90.7\%---an error rate a compliance office would not accept.

The dominant practice puts the entire SOP into the prompt as text. Resident prose (${\sim}10^4$ characters per turn on Bank) competes with the dialog for attention~\citep{liu2024lost}, and verbalization strips executable structure: alternative order, gate short-circuits, the boundary between \emph{what must hold} and \emph{how to verify it}. The symptoms: skipped checks, premature actions, incorrect refusals.

We treat the SOP as a program rather than a text. An offline, deterministic compiler translates the benchmark's machine-readable dependency constraints---the same source its verbalizer renders as the baseline prose---into a two-layer pseudo-code program: process functions, one per user goal, and rule subroutines pairing each constraint with its verification recipe and an evidence-bearing return. An online \emph{program-guided} (PG) runtime executes it as a two-layer virtual machine: a symbolic stack machine does the bookkeeping (stack, cursor, variables, recovery) and pages \emph{only the active frame} into context, while the LLM converses, calls tools, and judges evidence. Enforcement is deliberately \emph{soft}---attention, not permissions---at the price that correctness depends on faithful local-frame execution: a design bet we measure.

A three-arm design (official text; compiled text; compiled program plus runtime) separates the two layers on identical tasks, and six models spanning capability tiers chart where each layer pays. The compiled representation never significantly hurts and pays where official prose underperforms; a content$\times$format factorial shows the code-format package is likewise model-dependent. Runtime guidance divides more sharply: two strong models independently improve across seven domains, while weak models lose 14--26 points on Bank. A full-program cursor control localizes the strong-model benefit: making the active frame salient while retaining the global program recovers most of the refusal gain, and selective visibility adds a smaller increment. Paired probe-and-audit measurements trace the capability gate to spontaneous state discipline, corroborated at the attention level.

Our contributions are: (1) an SOP-to-program compiler and stack-paged runtime that make procedures executable and auditable; (2) a six-model attribution study whose seven-domain matrix independently reproduces positive PG effects for two strong models (58:19 and 75:31) and negative effects below the observed capability gate; (3) mechanism controls separating generic bookkeeping, active-frame salience, and selective visibility, alongside reconstruction, natural-audit, and attention diagnostics; and (4) a deployment rule---compile first, then page only after a model-level discipline check---plus a benchmark defect report.

\begin{figure*}[t]
\centering
\includegraphics[width=\textwidth]{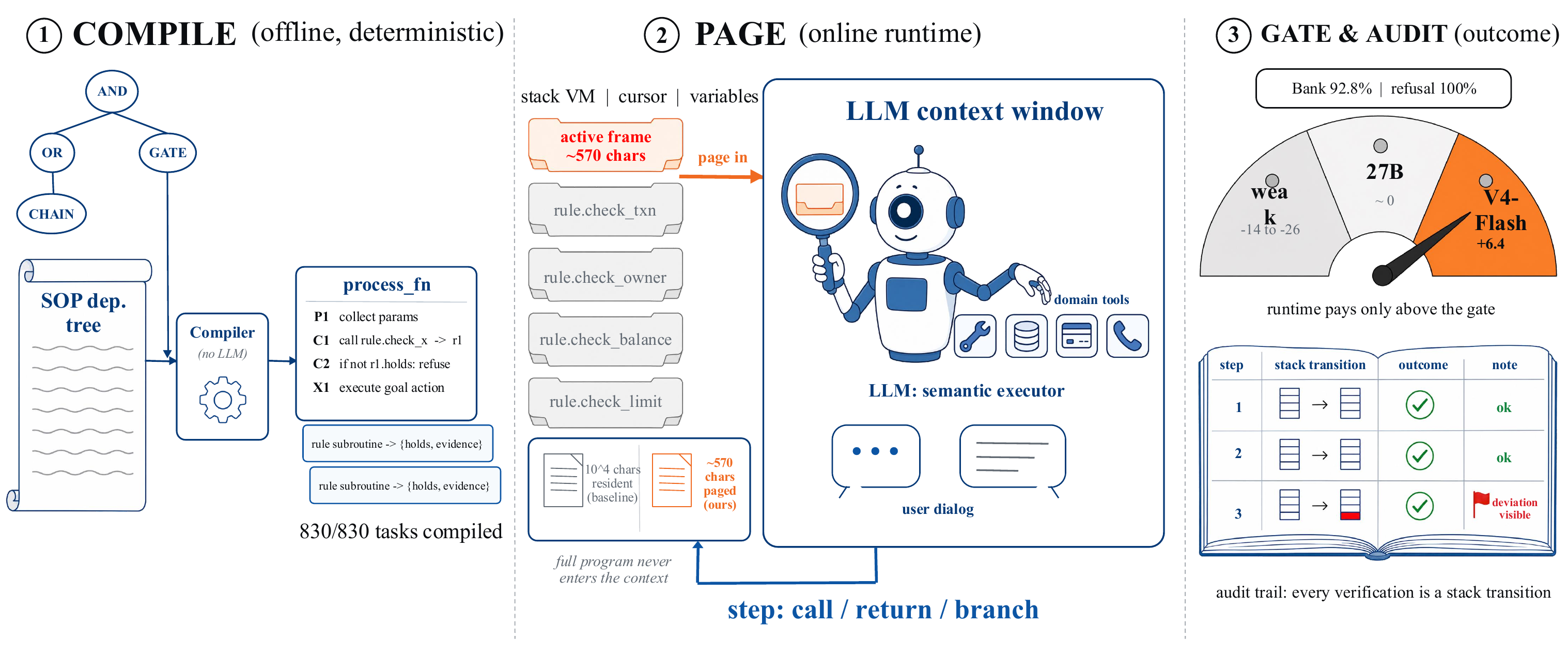}
\caption{Overview. (1)~An offline, LLM-free compiler turns the SOP dependency tree into a two-layer program (process functions calling rule subroutines that return $\{holds, evidence\}$; all 830 tasks compile). (2)~A stack VM pages \emph{only the active frame} (${\sim}570$ vs.\ ${\sim}10^4$ chars resident) while the LLM executes it semantically via one bookkeeping tool. (3)~Benefits are capability-gated; every verification is an auditable stack transition; deviations execute (soft enforcement) but are flagged.}
\label{fig:overview}
\end{figure*}

\section{Related Work}
\label{sec:related}

\paragraph{Progressive disclosure and context engineering.}
Managing what enters an agent's context is an established lever: resident long text is not reliably used~\citep{liu2024lost}, virtual context managers page \emph{memory} in and out~\citep{packer2023memgpt}, and Agent Skills~\citep{anthropic2025skills} disclose capability documents progressively at the \emph{document} level. InfiAgent externalizes persistent agent state to files~\citep{yu2026infiagent}; we use its public runtime substrate but add policy compilation and active-frame control. Our runtime pages \emph{policy} rather than memory, at the \emph{instruction} level, with call-graph-following, stateful, revocable disclosure (Method).

\paragraph{Programs that control language models.}
DSPy~\citep{khattab2024dspy}, LMQL~\citep{beurer2023lmql}, and SGLang~\citep{zheng2024sglang} compile or script \emph{pipelines of LM calls}: the program lives outside the model and invokes it. Finite-state scaffolds are an older form: task-oriented dialog systems pair a symbolic dialog manager with a statistical language layer~\citep{young2013pomdp}; StateFlow~\citep{wu2024stateflow} routes an LLM through hand-designed states. We invert this relationship: the program is the artifact the model must follow, and a symbolic runtime feeds it to the model frame-by-frame while the model remains the semantic executor inside the program.

\paragraph{SOP-following agents and hard enforcement.}
SOP-driven agents encode procedures as prompts or role scripts~\citep{sopagent2024,hong2024metagpt}, inheriting the resident-text failure modes quantified by our offtext arm; FlowAgent~\citep{shi2025flowagent} adds decision controllers over a procedure language, versus our stack-paged compiled program. At the opposite extreme, constrained decoding and FSM-style controllers~\citep{willard2023outlines} enforce structure \emph{hard}; this suits output formats, but procedures unfold over dialog turns where users change goals and evidence arrives in several valid orders, and a wrongly blocked action is unrecoverable. Our soft enforcement occupies the middle: every deviation remains possible but observable in the stack audit trail, and the cost of this flexibility---dependence on local instruction-following---is exactly the capability gate we characterize.

\paragraph{Agent benchmarks.}
General agent benchmarks measure tool competence~\citep{liu2024agentbench,yao2023react} or user-interaction reliability~\citep{yao2024taubench}; SOPBench~\citep{sopbench2025} grades against an oracle dependency graph with dual execute/refuse task types; we also report two confirmed simulator defects and three transparently screened omissions.

\section{Method}
\label{sec:method}
\begin{figure*}[t]
\centering
\includegraphics{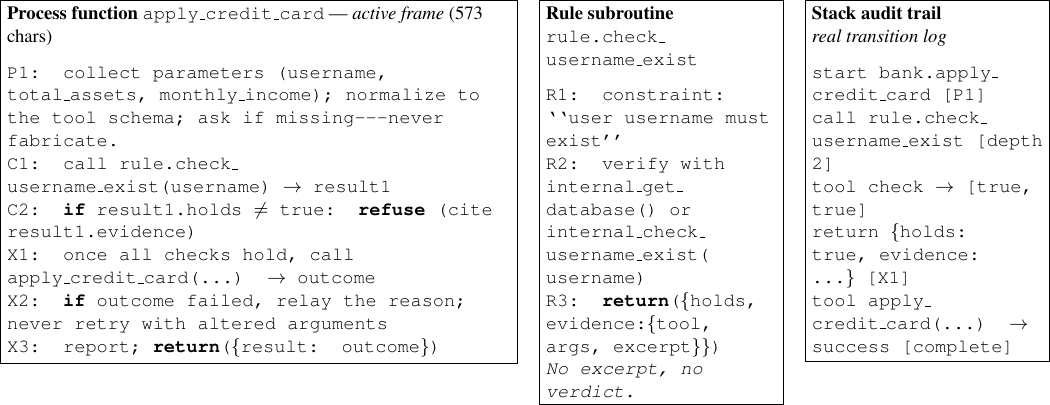}
\caption{A real compiled artifact (Bank task
\texttt{apply\_credit\_card\#0}). The original's Chinese structural
scaffolding (Setup) is rendered line by line in English---structure, labels,
identifiers verbatim. The runtime pages only the process frame (left) into
context; its verification becomes an evidence-returning subroutine (middle)
and an auditable stack transition (right).}
\label{fig:artifact}
\end{figure*}

We consider an agent that must complete user requests in a tool-augmented environment while complying with an SOP. Our approach replaces the dominant ``SOP as prompt text'' practice with a two-stage design: an offline \emph{compiler} turns the SOP into an executable pseudo-code program, and an online \emph{program-guided (PG) runtime} executes that program as a two-layer virtual machine, in which a symbolic runtime performs syntactic bookkeeping while the LLM acts as the semantic executor. The final subsection explains why enforcement is \emph{soft}; this choice preserves flexibility, and it predicts the capability threshold that Results observes. Figure~\ref{fig:overview} gives an overview.

% -------------------------------------------------------------------
\subsection{Compiling SOPs into Executable Programs}
\label{sec:compiler}

The official baseline verbalizes the same dependency trees used by our compiler, but drops alternative order, gate short-circuits, and the separation between what must hold and how to verify it. The compiler instead emits a two-layer program: one \emph{process function} per user goal calls \emph{rule subroutines} that pair the official constraint wording with a concrete verification recipe and return $\{\mathit{holds},\mathit{evidence}\}$. A false required result branches to \texttt{fail} before the goal action. Figure~\ref{fig:artifact} shows a real artifact and audit trail. Five mechanisms make the transformation executable rather than cosmetic:
\begin{enumerate}
\item \textbf{Stateful subroutines.} Interaction-requiring constraints compile to multi-turn functions whose progress persists across turns.
\item \textbf{Transitive verifier-gate hoisting.} Login/authentication gates required by verification actions are closed transitively and placed before guarded checks.
\item \textbf{Two disjunctions.} We adopt an evidence-complete schedule for policy \texttt{or}; this author-side convention preserves the benchmark's Boolean truth condition but is not required by its oracle. Alternative tool recipes are semantically equivalent paths and are robustness-ranked (full-state reads first).
\item \textbf{Frame factoring.} Large compound groups and composite alternatives become named functions, keeping continuations explicit.
\item \textbf{Evidence-bearing returns.} Semantic identifiers and explicit return contracts keep every verdict tied to its verifier.
\end{enumerate}

\paragraph{Formalization and guarantees.}
A leaf is a literal $\ell=(c,\theta,q)$: predicate $c$, parameter binding $\theta$, and required polarity $q\in\{0,1\}$ ($q=0$ denotes \texttt{not}~$c$). A rule is polarity-neutral: it returns $r=(h,e)$, where $h$ is the observed truth of $c(\theta)$ and $e$ is its tool evidence; $\ell$ holds iff $h=q$. For any evidence-faithful verdict assignment $\nu$,
\begin{align}
[\![\ell]\!]_\nu &= \mathbf 1\{\nu(c,\theta)=q\}, \nonumber\\[-2pt]
[\![\texttt{and}(\mathbf T)]\!]_\nu
=[\![\texttt{chain}(\mathbf T)]\!]_\nu
 &= \bigwedge_i[\![T_i]\!]_\nu, \nonumber\\[-2pt]
[\![\texttt{or}(\mathbf T)]\!]_\nu
=[\![\texttt{gate}(\mathbf T)]\!]_\nu
 &= \bigvee_i[\![T_i]\!]_\nu .
\label{eq:sop}
\end{align}
Their truth conditions coincide in pairs, but their schedules do not: \texttt{chain} preserves source order; under our evidence-complete convention ordinary \texttt{or} evaluates every option before deciding; \texttt{gate} tests options in source order and stops at the first success.

Let $A_R(T)$ be the verifier actions referenced by $T$, $\Gamma(A)$ the login/authentication gates required by actions $A$, and $A_G(X)$ the actions that establish gates $X$. The hoisted guard set is the least fixed point
\begin{equation}
\begin{aligned}
H_T&=\operatorname{lfp}_{X}\,\Gamma\!\left(A_R(T)\cup A_G(X)\right),\\[-2pt]
\widehat T&=\texttt{chain}\!\left(\operatorname{topo}(H_T\!\setminus L(T)),T\right).
\end{aligned}
\label{eq:closure}
\end{equation}
where $L(T)$ is the set of leaves already present in $T$; duplicates are removed. Up to semantics-preserving inlining, reuse, and frame factoring, compilation is
\begin{align}
\mathcal C(\ell)&=\mathsf{Check}(\ell), \nonumber\\[-3pt]
\mathcal C(\texttt{and}(\mathbf T))&=\mathsf{All}(\mathcal C(\mathbf T)), \nonumber\\[-3pt]
\mathcal C(\texttt{chain}(\mathbf T))&=\mathsf{All}^{\rightarrow}(\mathcal C(\mathbf T)), \nonumber\\[-3pt]
\mathcal C(\texttt{or}(\mathbf T))&=\mathsf{Any}^{\mathrm{all}}(\mathcal C^{\!*}(\mathbf T)), \nonumber\\[-3pt]
\mathcal C(\texttt{gate}(\mathbf T))&=\mathsf{Any}^{\mathrm{first}}(\mathcal C(\mathbf T)).
\label{eq:compile}
\end{align}
Here $\mathcal C^{\!*}$ is evidence-complete but truth-equivalent: $\mathsf{Any}^{\mathrm{all}}$ disjoins after all options return, whereas $\mathsf{Any}^{\mathrm{first}}$ short-circuits.

\noindent\textbf{Proposition 1} (Semantic and schedule preservation)\textbf{.}\label{prop:preserve}
\emph{For a conforming execution whose rule returns are faithful to their cited tool results, $\mathcal C(\widehat T)$ returns $[\![\widehat T]\!]_\nu$. Hence its prescribed control flow reaches the goal action iff the augmented SOP holds, and otherwise reaches \texttt{fail} first. Its trace preserves chain order, exhaustive ordinary-\texttt{or} evidence collection, and ordered \texttt{gate} short-circuiting, modulo same-frame reuse of an identical $(\mathrm{rule},\theta)$ result; it emits no predicate outside $L(T)\cup H_T$.}
The proof is a structural induction on Eq.~\eqref{eq:compile}; full cases and the fixed-point argument are in the supplement.

\paragraph{Technical advantages.}
All 830 artifacts load with closed call graphs and zero missing recipes. Compilation keeps the official leaf wording but adds recipes, guard closure, and evidence discipline; the content$\times$format factorial isolates these additions. It also reduces the average actionable instruction surface from ${\sim}10^4$ resident characters to a ${\sim}570$-character frame.

% -------------------------------------------------------------------
\subsection{The Program-Guided Runtime}
\label{sec:runtime}

A monolithic compiled program still competes with the dialog and does not externally preserve position. The runtime therefore executes it as a stack machine driven through one tool, \texttt{pg\_program\_step}, with \texttt{start}, \texttt{call}, \texttt{return}, \texttt{select\_branch}, \texttt{emit\_artifact}, \texttt{complete}, and \texttt{fail}. It maintains a call stack, cursor, and returned-variable store, and discloses only the active frame into a task-local context slot. A \texttt{call} reveals the callee; a \texttt{return} restores the suspended caller with the result bound and its cursor advanced. State persists outside the dialog, so the frame can be re-disclosed after interruption or context compression. The LLM remains the semantic executor: it converses, chooses tool arguments, interprets evidence, and reports verdicts through \texttt{return}.

\paragraph{Runtime semantics.}
A frame is $\phi=\langle f,\kappa,\eta,\rho\rangle$ (function, cursor, locals, caller return slot), and a configuration is $\sigma=(S,\Lambda)$ (frame stack and append-only trace). With the stack top on the right,
\begin{align}
S\!\cdot\!\phi
&\xrightarrow{\texttt{call }g(x)\to z}
S\!\cdot\!\phi\!\cdot\!\langle g,0,x,z\rangle, \nonumber\\[-2pt]
S\!\cdot\!\phi\!\cdot\!\langle g,\kappa',\eta',z\rangle
&\xrightarrow{\texttt{return }v}
S\!\cdot\!\operatorname{resume}(\phi,z,v).
\label{eq:vm}
\end{align}
Here $\operatorname{resume}(\langle f,\kappa,\eta,\rho\rangle,z,v)
=\langle f,\operatorname{next}_f(\kappa),\eta[z\leftarrow v],\rho\rangle$, and each transition is appended to $\Lambda$. Thus nested calls preserve pending continuations; lexical paragraph retrieval alone is insufficient unless it also reconstructs equivalent continuation state. Define
\begin{equation}
R(\sigma)=\operatorname{render}(\operatorname{top}(S)),\qquad
B_P=\max_{f\in F(P)}|\operatorname{render}(f)|.
\label{eq:reveal}
\end{equation}
The runtime injects only $R(\sigma)$, so $|R(\sigma)|\le B_P$, independent of call depth and dialog length (Bank mean: ${\sim}570$ vs. ${\sim}10^4$ resident characters).

\noindent\textbf{Proposition 2} (Well-bracketed auditability)\textbf{.}\label{prop:audit}
\emph{Every bound result has one dynamic producing \texttt{return}; all of its admissible evidence lies in the corresponding call--return interval. Consequently, a return without matching evidence, an out-of-cursor domain call, or a goal action before the accepting root cursor is detectable from the trace, although the runtime does not block it.}
This follows immediately from the well-bracketed transitions in Eq.~\eqref{eq:vm}; one producer may serve multiple same-frame consumers through explicit reuse.

\paragraph{Technical advantages.}
Paging is call-graph-following, stateful, and revocable: a return removes the callee from scope and restores the caller with bound results. On Bank it reduces domain calls from 3.0 to 2.8 and raises refusal correctness to 100\% for the strong workhorse (Results).

% -------------------------------------------------------------------
\subsection{Soft Enforcement and Bounded Discretion}
\label{sec:soft}

Hard blocking is brittle when goals change or evidence arrives in several valid orders. Our runtime therefore manages \emph{attention}, not permissions: every action remains possible, while the stack makes skipped checks and unsupported returns observable. This flexibility predicts a capability threshold---models must faithfully execute the local frame---which Results tests directly. The compiler is deterministic and LLM-free (${\sim}0.1$s for 830 programs); the runtime adds one bookkeeping tool and persists state per task.

\section{Experimental Setup}
\label{sec:setup}

\paragraph{Benchmark and clean subset.}
We evaluate on SOPBench~\citep{sopbench2025}, which casts SOP compliance as tool-use tasks over domain simulators with machine-readable dependency constraints and an oracle evaluator that checks both the executed call sequence (against a directed action graph) and the final database state. Bank is the primary attribution domain: 134 instances over 14 goals---\emph{execute} tasks (preconditions satisfiable; act) and \emph{refusal} tasks (unsatisfiable; refuse). Before comparing arms, an upstream zero-pass screen omitted 9 execute tasks from three Bank goals; current-source replay confirms simulator/state-transition defects for two, whereas \texttt{open\_account} is an empirical exclusion rather than a proved impossibility (supplement). We report the resulting screened subset of 125 (39 execute, 86 refusal), and recompute official baselines on it. Conservative full-denominator scores treating all nine omissions as failures are offtext/flat/\pg{}: 65.7/80.6/86.6 over 134; paired statistics are unchanged because the omissions are arm-independent. The runtime contrast additionally runs on the other six domains (Table~\ref{tab:matrix}) with per-domain budgets (deep domains: 1800\,s, 5 turns). The same pre-run screen omitted one upstream-zero-pass execute task in each of Library and DMV; current-source replay finds both reachable, so we do not call them confirmed defects.

\paragraph{Three arms: isolating representation from runtime.}
All arms share the agent loop, domain tools, simulator, and budgets, differing only in how the SOP reaches the model: \textbf{offtext} is the benchmark's official verbalized constraint text, resident every turn (the SOP-as-prompt baseline); \textbf{flat} is the full text of our compiled program, resident every turn (compiled \emph{representation} without the runtime); \textbf{\pg{}} is the same compiled program executed by the PG runtime with just-in-time instruction paging (Method).
The offtext$\rightarrow$flat delta measures \emph{compilation} and flat$\rightarrow$\pg{} the \emph{runtime} effect. A diagnostic \textbf{full-cursor} arm uses the same cursor-guided runtime and places the active frame first while retaining the exact full compiled program below it as global reference; it bundles frame repetition/position, cursor guidance, and global visibility rather than isolating pure highlighting. Its V4 three-arm common set has $n{=}819$ across seven domains. All tests are paired (exact McNemar; exact binomial pooling; per-contrast, unadjusted), with no post-hoc adherence filtering.
\paragraph{Models.}
Six assistant models span capability tiers: DeepSeek-V4-Flash (low-cost workhorse) and Qwen3.7-Plus (strong), GPT-4o-mini and Qwen2.5-7B-Instruct (weak), plus a same-vendor probe pair (Qwen3.6-27B/-35B-A3B) added to resolve the transition band. The user simulator is DeepSeek-V4-Flash for all arms and models. Three diagnostic arms used only in the compilation analysis are defined in Results.

\paragraph{Agent loop and budgets.}
Every task runs with identical caps: 150 agent turns, an attention-window refresh every 50 actions, and fuses of 900\,s wall clock, 30 domain calls, 3 user turns. Each run gets a fresh task identity; every record embeds a configuration fingerprint (all Bank results share one); outcomes are scored solely by the official evaluator.

\paragraph{Comparability disclosures.}
Relative to the official leaderboard, our simulator, user-turn budget (3 vs. ${\sim}$10), and clean task set differ; the within-harness offtext arm therefore anchors every causal contrast.  Sensitivity checks preserve the result structure: at the official turn budget, \pg{} remains above flat (90.4 vs. 87.2); a tied-top official model is slightly \emph{lower} in our offtext loop (79.2 vs. 82.4 official); and a cross-family simulator reproduces the three-arm ordering (72.8/83.2/89.6).  Compiled arms use Chinese structural scaffolding with English leaves; English re-emissions reproduce flat at both endpoints, while the weak model's isolated code-format harm attenuates from $-12.8$ to $-5.6$ (n.s.).  Full protocols and discordants are in the supplement.

\section{Results}
\label{sec:results}
% 本文件由 paper/scripts/gen_tables.py 生成，勿手改

\begin{table}[t]
\centering
\small
\setlength{\tabcolsep}{2pt}
\begin{tabular*}{\columnwidth}{@{\extracolsep{\fill}}lccccc@{}}
\toprule
Model & offtext & flat & \pg{} & flat:off ($p$) & \pg{}:flat ($p$) \\
\midrule
DS-V4-Flash & 70.4 & 86.4 & 92.8 & 26:6 ($<$.001) & 9:1 (.021) \\
Q3.7-Plus & 85.6 & 85.6 & 88.8 & 9:9 (1.000) & 6:2 (.289) \\
Q3.6-A3B & 81.6 & 81.6 & 84.0 & 14:14 (1.000) & 12:9 (.664) \\
Q3.6-27B & 79.2 & 86.4 & 85.6 & 16:7 (.093) & 4:5 (1.000) \\
Q2.5-7B & 65.6 & 68.8 & 54.4 & 22:18 (.636) & 14:32 (.011) \\
GPT-4o-mini & 60.8 & 75.2 & 48.8 & 32:14 (.011) & 9:42 ($<$.001) \\
\bottomrule
\end{tabular*}
\caption{Three arms across six models (Bank clean subset, one harness, $n{=}125$ per arm; all cells pass\,\%). Right columns: discordant pairs with exact McNemar $p$ (compilation flat:offtext; runtime \pg{}:flat); point effects are column differences. 27B/A3B: same-vendor transition probe pair; full names in Setup.}
\label{tab:models}
\end{table}

\begin{table}[t]
\centering
\small
\begin{tabular*}{\columnwidth}{@{\extracolsep{\fill}}lcc@{}}
\toprule
Official configuration & Pass\,\% & Refusal\,\% \\
\midrule
o4-mini-high (FC) & 82.4 & 90.7 \\
GPT-4o (FC) & 82.4 & 84.9 \\
Qwen2.5-32B (ReAct) & 82.4 & 90.7 \\
Qwen2.5-7B (FC)$^\dagger$ & 5.2 & -- \\
\bottomrule
\end{tabular*}
\caption{Top official SOPBench configurations, recomputed on the identical clean subset (the three tied-top configurations; the remaining fifteen range 70.4--80.8). $^\dagger$Official full-set leaderboard score (refusal split unavailable): the 7B cross-harness reference.}
\label{tab:official}
\end{table}

\begin{table}[t]
\centering
\small
\setlength{\tabcolsep}{2pt}
\begin{tabular*}{\columnwidth}{@{\extracolsep{\fill}}lccccc@{}}
\toprule
Model & offtext & off-code & c-prose & $\Delta_{\mathrm{fmt}}$ & $\Delta_{\mathrm{cont}}$ \\
\midrule
DS-V4-Flash & 70.4 & 84.8 & 88.0 & 26:8 (.003) & 27:5 ($<$.001) \\
Q2.5-7B & 65.6 & 52.8 & 68.0 & 10:26 (.011) & 21:18 (.749) \\
\bottomrule
\end{tabular*}
\caption{Compilation $2{\times}2$: \{official, compiled\} content $\times$ \{prose, code\} format, offtext as reference. \emph{off-code} re-syntaxes official text with byte-identical leaves (machine-checked); \emph{c-prose} renders compiled semantics as prose. $\Delta_{\mathrm{fmt}}$/$\Delta_{\mathrm{cont}}$ vs.\ offtext; both carry the scaffold-language change (Setup); flat column in Table~\ref{tab:models}.}
\label{tab:p1}
\end{table}

\begin{table}[!t]
\centering
\small
\setlength{\tabcolsep}{0.8pt}
\begin{tabular*}{\columnwidth}{@{\extracolsep{\fill}}lcccc@{}}
\toprule
Domain & V4 & Plus & 27B & 7B \\
\midrule
Bank & 9:1 & 6:2 & 4:5 & 14:32 \\
Hotel & 13:7 & 26:16 & -- & -- \\
Library & 2:1 & 3:4 & 9:6 & 5:19 \\
DMV & 4:1 & 5:3 & 0:4 & 10:43 \\
Healthcare & 16:6 & 20:3 & 18:3 & 14:54 \\
Market & 14:2 & 15:3 & 9:11 & 15:66 \\
University & 0:1 & 0:0 & 1:1 & 1:23 \\
\midrule
Pooled W:L & 58:19 & 75:31 & 41:30 & 59:237 \\
Pooled $p$ & $<$.001 & $<$.001 & .235 & $<$.001 \\
\bottomrule
\end{tabular*}
\caption{Runtime contrast across seven domains: \pg{}:flat discordant pairs (wins:losses for \pg{}). The two strong models are evaluated across all domains; transition/weak cells follow the screened sets in Setup. Pooled rows use exact McNemar tests.}
\label{tab:matrix}
\end{table}

\begin{table}[t]
\centering
\small
\setlength{\tabcolsep}{0pt}
\begin{tabular*}{\columnwidth}{@{\extracolsep{\fill}}lcccc@{}}
\toprule
Split & $n$ & flat & full-cursor & \pg{} \\
\midrule
All & 819 & 88.2 & 91.0 & 92.9 \\
Execute & 277 & 91.7 & 90.3 & 91.0 \\
Refusal & 542 & 86.3 & 91.3 & 93.9 \\
\bottomrule
\end{tabular*}
\caption{Full-program cursor ablation on the V4 three-arm common set (pass\,\%). Full-cursor keeps the active frame first and the complete compiled program below it. Overall discordants: full-cursor:flat 57:34 ($p{=}.021$); \pg{}:full-cursor 36:20 ($p{=}.044$).}
\label{tab:fullcursor}
\end{table}

% Generated by paper/scripts/gen_figures.py; do not edit manually.
\begin{figure}[t]
\centering
\includegraphics{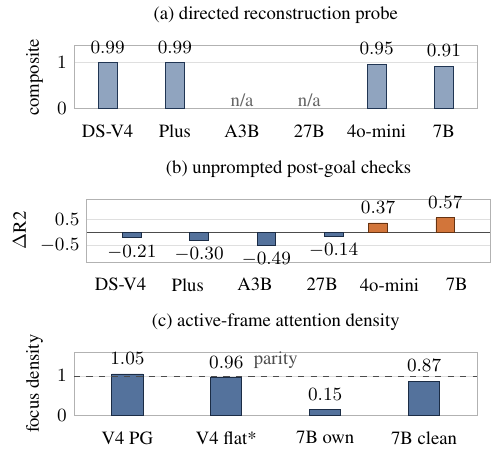}
\caption{Mechanism. (a)~Directed probes: near ceiling for all four models measured (transition pair not probed). (b)~Wild audit ($\Delta$R2 $=$ excess post-goal calls, \pg{}$-$flat): the two harmed models are the two positive bars; the four unharmed are negative. (c)~Attention over length share (1.0$=$parity): V4 at parity on its 2.1\%-of-context frame; the 7B under-allocates to it on its own trajectories yet nears parity re-presented cleanly. *flat's region is the whole program.}
\label{fig:mech}
\end{figure}

\subsection{Attribution: Compilation vs.\ Runtime}
\label{sec:attribution}

The first row of Table~\ref{tab:models} decomposes the end-to-end gain on Bank (clean subset, DeepSeek-V4-Flash) into its two layers.
Replacing the official SOP text with the \emph{compiled program text} --- same channel, no runtime --- lifts pass rate from 70.4\% to 86.4\% (+16.0 points; 26:6, $p<0.001$).
Enabling the \pg{} runtime on the same program adds +6.4 within our harness, reaching 92.8\% on the clean subset (86.6\% on the full 134-task set; 9:1, $p{=}.021$), with 100\% refusal correctness (86/86; two-sided 95\% lower bound 95.8\%; reproduced in both re-runs), fewer domain calls (3.0$\rightarrow$2.8), and wall clock 181/105/132\,s. Our PG configuration exceeds every recomputed official configuration on both metrics (Table~\ref{tab:official}); we treat this as a qualified external reference, since simulator, turn budget, and loop differ (Setup).
The execute/refusal decomposition: execution moves little (74.4$\rightarrow$71.8$\rightarrow$76.9) while refusal climbs 68.6$\rightarrow$93.0$\rightarrow$100---what compilation and paging chiefly buy a strong model is \emph{correct refusal behavior}.
The offtext column of Table~\ref{tab:models} extends this across six models: flat is never significantly below offtext, and the compilation gain lands where official prose underperforms (+16.0/+14.4 significant; +7.2/+3.2 n.s.), vanishing where official prose already sits at flat level (85.6, 81.6).

\subsection{The Runtime Gain is Capability-Gated}
\label{sec:threshold}

Across six models (Table~\ref{tab:models}), the runtime effect $\Delta_{\mathrm{rt}}$ is not noise around zero: it spans a signed range from $+6.4$ (DeepSeek-V4-Flash, $p{=}0.021$) and $+3.2$ (n.s.), through a same-vendor probe pair on opposite sides of zero ($+2.4$/$-0.8$, both n.s.), down to $-14.4$ ($p{=}0.011$) and $-26.4$ (GPT-4o-mini, $p{<}0.001$); we claim no independent capability ordering.
Pooling the two strong models yields 15:3 discordants; a task-level permutation test preserving the cross-model correlation (the models share tasks) gives $p{=}0.016$; the two weak models are independently significantly negative.
Benjamini--Hochberg across the twelve per-model contrasts retains both compilation gains and both weak-model harms but not the single-model +6.4---runtime claims rest on the pooled discordants and replicate signs.
Because instances within a goal share a program template, we also re-test with a goal-clustered sign-flip permutation (all discordants of a goal flip together): every cross-domain claim survives clustering (V4 58:19 and Plus 75:31, both $p{=}.004$; weak 59:237 at $p{\le}10^{-4}$; GPT-4o-mini's Bank harm at $p{=}.028$), while the Bank-only strong-pair 15:3 spans just seven goal clusters and is underpowered there ($p{=}.19$)---a further reason runtime claims rest on cross-domain pools. Goal aggregates are in the supplement and complete per-goal results in the artifact.
Identical-configuration re-runs keep every model on the same side of zero: the strong gain recurs, transition effects remain near zero, and the severe weak-model harm recurs at $-24.0$ to $-33.6$ (all $p{<}0.001$); full runs are in the supplement.
A decoupling arm then attributes the effect \emph{within} the runtime: \emph{flat+protocol} keeps the full program resident while mandating the same bookkeeping protocol---paging off.
Its overhead is insignificant at all three tiers (V4-Flash 3:4; GPT-4o-mini 11:20; 7B 22:15; all n.s.), ruling out generic bookkeeping as the primary cause. Table~\ref{tab:fullcursor} then retains the complete program but places the active frame first. On the V4 common set, flat/full-cursor/\pg{} score 88.2/91.0/92.9: full-cursor beats flat 57:34 ($p{=}.021$), while \pg{} further beats full-cursor 36:20 ($p{=}.044$). The separation is refusal-centered (86.3/91.3/93.9); execution is unchanged. Thus active-frame salience recovers much of the strong-model gain without hiding global instructions, and selective visibility adds a smaller increment.

The weak-model boundary is different: on Bank, 7B flat/full-cursor/\pg{} score 68.8/55.2/54.4, and full-cursor is indistinguishable from \pg{} (21:20, $p{=}1$). Restoring the full program therefore does not recover its loss, suggesting that active-frame-guided execution itself can exceed the model's usable control capacity. GPT-4o-mini instead \emph{over-verifies}, continuing checks after the goal action.
Quantifying: 19 of its 64 \pg{} failures (10 of 7B's 57) are \emph{outcome-correct} (right final state, no violation), failing only the call-sequence check---absent under flat for every model; crediting all such cases still leaves it 11.2 below flat (64.0\%): the gate is not a grading artifact.

\subsection{Seven Domains across Capability Tiers}
\label{sec:hotel}

Table~\ref{tab:matrix} extends the runtime contrast to all seven domains for two strong models, with transition and weak models on the five new domains.
V4 pools to 58:19 ($p{\approx}10^{-5}$), with six domains leaning \pg{}; the independent Plus replication pools to 75:31 over 821 pairs ($p{=}2.3\!\times\!10^{-5}$), with five domains leaning \pg{}, Library 3:4, and University tied. This directly removes the single-strong-model dependence of the cross-domain finding without claiming a universal numerical threshold. The transition model pools to 41:30 (n.s.), splitting by domain; the weak model is harmed throughout, pooling to 59:237 ($p{\approx}10^{-26}$; disjoint tasks, exact pools).
No cell reverses its tier's sign significantly.
Healthcare rewards PG for both strong endpoints and the transition model (V4 16:6; Plus 20:3; 27B 18:3)---long but strictly tool-checkable chains.
Deep-program domains (Hotel, up to 14 functions; University) show at most a modest strong-tier edge: on Hotel, \pg{} passes 92.3\% vs.\ flat 89.2\% (13:7, n.s.) with the gain concentrated on refusal correctness (96.0\% vs.\ 87.3\%) while flat leads execution by 7.3 points---deep-domain constraints \emph{look} inferable, inviting substitute-verification---at a wall-clock price (418 vs.\ 166\,s/task).

\subsection{Why Compilation Works: Content vs.\ Format}
\label{sec:p1}

Compilation both \emph{adds content} (verification recipes, verifier-gate closure, evidence discipline) and \emph{re-formats} the policy as code.
Table~\ref{tab:p1} separates the two with a $2{\times}2$ over \{official, compiled\} content $\times$ \{prose, code\}, anchored to the shared offtext baseline, using two equivalence-preserving transforms (\emph{off-code}: official text re-syntaxed with byte-identical leaves; \emph{c-prose}: compiled semantics as prose, same checks, order, gates, recipes; shared scaffolding conventions); the language ablations of Experimental Setup detect no significant language effect, and the off-code cell is measured directly: in English the weak model's format harm attenuates to $-5.6$ (n.s.), direction unchanged.

Both factors are capability-dependent, in different ways: format flips sign, content pays only above a capability line.
The \emph{format} effect at fixed content flips sign across the capability range: re-syntaxing the official prose into code gains the strong model +14.4 points (26:8, $p{=}0.003$) yet costs the weak model $-12.8$ (10:26, $p{=}0.011$).
The \emph{content} effect at fixed prose format is +17.6 for the strong model (27:5, $p{<}0.001$) but not detected for the weak one (+2.4, 21:18, $p{=}.75$); once content is compiled, format no longer matters (flat vs.\ c-prose: $-1.6$/$+0.8$, both n.s.).
For the strong model, mechanical structure explicitation alone thus recovers most of compilation's value; for the weak model, no representation manipulation significantly \emph{improves} it within our harness (code re-syntaxing significantly harms it)---its dramatic lift over the official leaderboard (official full set 5.2\%, ${\approx}5.6\%$ rescaled, vs.\ 65.6\% at our offtext arm) is attributable to the execution harness, not the representation.
Practically: the compiled \emph{representation} never significantly hurt; isolated \emph{content} pays above a low capability line; code \emph{syntax} should be reserved for models it demonstrably helps.

A natural objection---the content effect merely leaks verification recipes---fails an ablation: appending complete recipes to the \emph{unmodified} official text recovers a minority (70.4$\rightarrow$76.8, n.s.), while compiled arms stay significantly ahead (flat +9.6, $p{=}.029$; c-prose +11.2, $p{=}.016$).
Compilation buys not the recipe \emph{information} but its \emph{organization}: obligations, recipes, and evidence discipline in one goal-scoped executable flow.

\subsection{Why the Runtime is Gated: Discipline, not Reconstruction}
\label{sec:mechanism}

We probed the threshold with paired behavioral and white-box measurements (Figure~\ref{fig:mech}).

\paragraph{Mechanism: capability present, deployment divergent.}
A directed state-reconstruction probe (machine-graded against the program cursor) scores \emph{all} four probed models near ceiling (0.91--0.99, localization 100\%), including both runtime losers: the gate is not inability to reconstruct state when asked.

An unprompted audit --- R2, mean domain calls issued \emph{after} the goal action already succeeded (successful-goal trajectories; no probe asked) --- separates harmed from unharmed in the initial runs with a single sign rule, no exemptions: the two significantly harmed models are exactly the two with positive signatures ($+0.37$, $+0.57$); all four unharmed carry negative ($-0.14$ to $-0.49$).
Both transition-band models, audited \emph{after} the rule froze, landed unharmed---a prospective check on fresh trajectories.
Across re-runs the rule separates \emph{harm} robustly but its null region is noisy: the harmed signature recurs in all three runs (GPT-4o-mini: $+0.37/+0.12/+0.48$), while the strong models' near-zero signatures drift positive in the second re-run ($+0.09$, $+0.15$)---a one-sided harm detector, not a two-sided predictor.
Reconstruction capability is present wherever probed; \emph{spontaneous} state discipline is what the runtime gate tracks.

Running the two open-weight endpoints on their own passing trajectories (architecture-faithful attention capture; real-token positions only):
DeepSeek-V4-Flash spontaneously grants its 2.1\%-of-context frame an at-parity share (1.05 vs.\ flat's 0.96), and ablating the entire dialog history \emph{raises} the gold action's log-probability by 1.9 nats: the frame alone is sufficient to identify the target action.
Qwen2.5-7B, in contrast, gives the frame only $0.15\times$ parity on its own trajectories yet $0.87$ re-presented cleanly: the capability--deployment gap at the attention level.

\section{Discussion}
\label{sec:discussion}

\paragraph{Deployment guidance.}
The results reject ``more scaffolding is better.'' Compile when official prose underperforms; use code syntax only after a model-level check and paging only after a discipline check. Strong models benefit from active-frame salience, with a smaller refusal-centered gain from selective visibility. For weak models, use the strong execution harness without active-frame guidance. Paging is soft and auditable, so irreversible actions still require hard tool-layer guards.

\paragraph{Limitations.}
(i) One benchmark (seven domains). (ii) Simulator/harness differences may leave residual affinity despite offtext, anchor, and turn-budget controls. (iii) Six models and no quantitative capability scale. (iv) Pass@1 without sampling-variance estimates; re-runs preserve each model's gate side. (v) Machine-readable trees only; natural-language input unevaluated. An authorized, de-identified industrial 13-function/87-step case completed an employee-run workflow---qualitative portability evidence only, not compiler validation or matched comparison. (vi) Decoupling separates protocol from paging, but not state externalization from disclosure. Artifacts will be released.

\section{Conclusion}
We presented a deterministic SOP compiler and stack-paged runtime separating representation from visibility. Across six models and seven domains, two strong models independently benefit while weak models can be harmed. Full-cursor results recover most of the strong-model refusal gain through active-frame salience; selective visibility adds a smaller increment. Compile first; page only after a model-level discipline check. AI disclosure: Claude aided coding; GPT polished author-written text; authors verified all outputs.

\bibliographystyle{plainnat}
\bibliography{refs}

\clearpage
\appendix
\section{Formal Details}
\label{app:formal}

\paragraph{Abstract execution.}
For a literal $\ell=(c,\theta,q)$, let a rule execution return
$r_\ell=(h,e)$, where $h\in\{0,1\}$ and $e$ is a set of logged tool
events.  A return is \emph{evidence-faithful} when the cited event uniquely
identifies the verifier action, its arguments agree with $\theta$, and its
observed output entails $h$.  The proofs below concern conforming executions:
the executor follows the generated call, return, and branch instructions, while
rule-level semantic judgments satisfy this evidence contract.  The production
runtime is deliberately soft and records, rather than blocks, violations of
this contract.

\paragraph{Guard closure.}
Let $\mathcal G$ be the finite set of login/authentication predicates.  For an
SOP tree $T$, define the monotone operator
\begin{equation}
F_T(X)=\Gamma\!\left(A_R(T)\cup A_G(X)\right),\qquad X\subseteq\mathcal G.
\end{equation}
Because $\mathcal G$ is finite, the ascending sequence $X_0=\varnothing$ and
$X_{k+1}=F_T(X_k)$ reaches the least fixed point $H_T$ in at most
$|\mathcal G|$ strict extensions.

\noindent\textbf{Lemma 1} (Minimal verifier-guard closure)\textbf{.}
\emph{$H_T$ contains every gate required by an action used to verify $T$ and,
recursively, every gate required to establish a gate already in $H_T$.
Moreover, every member of $H_T$ is justified by one of these two cases.}

\emph{Proof.}
The first statement follows from $H_T=F_T(H_T)$.  For minimality, let $Y$ be
any set satisfying the two closure conditions.  By induction,
$X_k\subseteq Y$ for all $k$: the base case is immediate, and monotonicity of
$F_T$ gives $X_{k+1}=F_T(X_k)\subseteq F_T(Y)\subseteq Y$.  Taking the stable
limit yields $H_T\subseteq Y$. \hfill$\square$

\paragraph{Evidence-complete disjunction.}
Write $\operatorname{val}_\nu(P)$ for the Boolean returned by an abstract
program $P$ under verdict assignment $\nu$.  The compiler uses two operational
forms of disjunction:
\begin{align}
\mathsf{Any}^{\mathrm{all}}(P_1,\ldots,P_m)
  &: \begin{aligned}[t]
     &\text{execute every $P_i$, then return}\\[-2pt]
     &\textstyle\bigvee_i\operatorname{val}_\nu(P_i),
     \end{aligned}\\
\mathsf{Any}^{\mathrm{first}}(P_1,\ldots,P_m)
  &: \begin{aligned}[t]
     &\text{execute in order and return}\\[-2pt]
     &\text{at the first true $P_i$}.
     \end{aligned}
\end{align}
The evidence-complete transform $\mathcal C^{\!*}$ suppresses premature
failure inside an ordinary-\texttt{or} option, accumulates the option's required
verdicts, and returns their Boolean composition only after collection.  Nested
\texttt{gate} nodes retain ordered short-circuit semantics.

\noindent\textbf{Lemma 2} (Evidence completion preserves truth)\textbf{.}
\emph{For every SOP subtree $T$,
$\operatorname{val}_\nu(\mathcal C^{\!*}(T))=[\![T]\!]_\nu$.
At each ordinary-\texttt{or} node, every child option is invoked before the
node returns; at each \texttt{gate}, no child after the first successful child
is invoked.}

\emph{Proof.}
By structural induction.  A leaf compares the polarity-neutral verdict with
its required polarity.  For \texttt{and} and \texttt{chain}, evidence-complete
mode evaluates the required children and returns their conjunction; source
order is retained for \texttt{chain}.  For ordinary \texttt{or}, the induction
hypothesis gives the correct value for each child, all children are invoked by
$\mathsf{Any}^{\mathrm{all}}$, and the returned value is their disjunction.
For \texttt{gate}, $\mathsf{Any}^{\mathrm{first}}$ returns true exactly when a
child is true and otherwise exhausts the list, which is Boolean disjunction
with the prescribed first-success schedule. \hfill$\square$

\paragraph{Proof of Proposition 1.}
By Lemma~1, $\widehat T$ adds exactly the minimal verifier guards not already
present in $T$, in prerequisite order.  We prove
$\operatorname{val}_\nu(\mathcal C(U))=[\![U]\!]_\nu$ for every subtree $U$
of $\widehat T$ by structural induction.
For a leaf, evidence faithfulness supplies the observed truth $h$ and
$\mathsf{Check}$ returns $\mathbf 1\{h=q\}$.  The \texttt{and} case follows
from conjunction of the induction hypotheses; the \texttt{chain} case is the
same while preserving source order.  For ordinary \texttt{or}, Lemma~2 and
$\mathsf{Any}^{\mathrm{all}}$ yield the disjunction only after every option
returns.  For \texttt{gate}, $\mathsf{Any}^{\mathrm{first}}$ implements the
same disjunction with ordered short-circuiting.  Extracting a subtree into a
named frame, inlining it, or reusing an identical same-frame result changes
neither the returned value nor the relative order of non-reused checks.
Finally, the process wrapper branches to \texttt{fail} on a false root result
and exposes the goal action only on a true root result.  The compiler emits
rules only while traversing $T$ or $H_T$, establishing the no-invention claim.
\hfill$\square$

\paragraph{Proof of Proposition 2.}
The only transition that binds a callee result into a caller is
\texttt{return}.  Stack discipline therefore assigns each dynamic binding a
unique producer and a unique innermost call--return interval; explicit
same-frame reuse may create multiple consumers but not another producer.
Since framework and domain calls are appended to the same trace, an evidence
reference can be checked for membership, tool identity, and argument agreement
inside that interval.  A domain call not licensed by the active cursor, or a
goal action before the root reaches its accepting cursor, is likewise visible
by comparing the event with the contemporaneous top frame.  These checks are
observational: soft enforcement leaves the violating action executable.
\hfill$\square$

\paragraph{Artifact validation.}
For all 830 generated artifacts, the released validator checks that the
production loader accepts the program, the entry function exists, every body
parses into steps, all declared and textual call targets resolve, and YAML
field types are valid.  This validates artifact well-formedness and call-graph
closure; semantic preservation is the construction-level result proved above,
not an empirical equivalence claim made by the loader test.

\section{Compiler Engineering and Artifact Audit}
\label{app:engineering}

\paragraph{What is automatic.}
The compiler reads the benchmark's goal schemas, dependency trees, constraint
verbalizations, and stateless predicate-to-tool verification recipes.  It has
no goal-specific mapping table.  The only hand-written stateful predicate
templates are \texttt{logged\_in\_user} and
\texttt{authenticated\_admin\_password}; they cover multi-turn login and
administrator authentication.  The remaining hand-written compiler policies
are domain agnostic: transitive gate closure, the two disjunction schedules,
frame factoring at a seven-leaf budget, identical-result reuse, and deterministic
robustness ordering of alternative verifier recipes.  Thus a new domain that
exposes the same metadata interface and no new stateful interaction type needs
no goal-level compiler edit.  A genuinely new stateful interaction type needs
one reusable subroutine template; a missing verifier recipe must be added to
the source metadata rather than patched per task.

\begin{table}[t]
\centering
\small
\begin{tabular*}{\columnwidth}{@{\extracolsep{\fill}}lr@{}}
\toprule
Quantity & Count \\
\midrule
Compiler source lines (nonblank/noncomment) & 1,118 (978) \\
Factorial/language emitters & 432 (372) \\
Production-loader validator & 112 (94) \\
Goal-specific hand mappings & 0 \\
Hand-written stateful predicate templates & 2 \\
Generated artifacts / unique goal entries & 830 / 70 \\
Unique rule-function identifiers & 165 \\
Functions per artifact, median / maximum & 7 / 21 \\
Artifacts using a missing-recipe fallback & 0 \\
\bottomrule
\end{tabular*}
\caption{Mechanical compiler-effort audit. Source counts include bilingual
templates and formatting code so that the engineering surface is explicit.}
\label{tab:compiler-audit}
\end{table}

The 830 artifacts comprise Bank 134, DMV 97, Healthcare 124, Hotel 195,
Library 66, Online Market 172, and University 42.  Their median serialized
length is 7,970 characters (range 1,500--26,604).  The upstream metadata
provides 17/17/19/32/17/24/37 verification recipes in these domains,
respectively.  The released audit script regenerates every count in
Table~\ref{tab:compiler-audit}.

\paragraph{Validation boundary.}
Running the production PG loader over the release gives 830/830 accepted
artifacts.  For each file the validator checks a declared entry, nonempty
parsed bodies, declared and textual call-target resolution, and YAML field
types.  This is a syntax, loader-compatibility, and call-graph-closure result.
It is not a task-by-task empirical proof that an LLM will return a faithful
predicate verdict; Proposition 1 states the separate construction-level claim
and makes that assumption explicit.

\section{Experimental Protocol and Reproduction Details}
\label{app:protocol}

\paragraph{Primary arms and scoring.}
All paired arms receive the same task instance, initial database, domain tools,
user simulator, and evaluator.  \emph{offtext} keeps the official verbalized
SOP resident; \emph{flat} keeps the complete compiled program resident; and
\pg{} exposes the same program through the stack runtime.  The official
evaluator requires both the directed action graph and final database state.
We use pass@1 with temperature 0, no outcome-conditioned retry, and a fresh
task identity for every run.  Infrastructure failures are discarded and
rerun under a new identity; every retained JSON records its run fingerprint.

\emph{Full-cursor} is a bundled visibility control: the ordinary active frame
and cursor guidance remain first, while the exact task-local compiled program
is appended below as global reference. It therefore combines active-frame
repetition/position, cursor guidance, and full-program visibility; we do not
call it a pure highlighting intervention. Its primary analysis uses all
assigned outcomes on the V4 flat/full-cursor/\pg{} common set ($n{=}819$),
without filtering on observed tool use or program markers. Full-cursor itself
completed all 821 eligible tasks. The Plus replication is analyzed on its
complete paired outcome set, without outcome-conditioned filtering.

The assistant endpoints are
\path{openrouter/deepseek/deepseek-v4-flash},
\path{openrouter/qwen/qwen3.7-plus},
\path{openrouter/qwen/qwen3.6-35b-a3b},
\path{openrouter/qwen/qwen3.6-27b},
\path{openrouter/qwen/qwen-2.5-7b-instruct}, and
\path{openrouter/openai/gpt-4o-mini}, accessed in July 2026.
The common configuration uses temperature 0, provider-default output cap,
128k configured context, structured tool calling, 600s request timeout, 150
agent turns, a 50-action refresh window, 30 domain calls, and 3 user turns.
Deep domains use 1,800s and 5 user turns; other domains use 900s and 3.
The user simulator is DeepSeek-V4-Flash. Task data determine simulator state;
there is no sampled task-order seed in a per-instance pass@1 comparison.

\paragraph{Compute and software.}
The API experiments were orchestrated on a commodity 10-core, 16GB host;
model inference was remote and no local accelerator affected the primary
outcomes.  The environment used Python
3.12.7, LiteLLM 1.82.6, PyYAML 6.0.3, NumPy 1.26.4, and SciPy 1.17.1.  The
open-weight attention diagnostics ran on two NVIDIA RTX PRO 6000 GPUs with
PyTorch 2.4.0 and Transformers 4.36.0.  The anonymous environment snapshot and
lockable requirements are included in the artifact.

\paragraph{Statistics.}
Within a model/domain we use exact McNemar tests on paired task outcomes.  A
pooled task analysis uses exact sign/binomial calculations, and the strong
two-model Bank test flips a task jointly across models.  Because instances
within a goal share program structure, we additionally flip all discordant
pairs belonging to the same goal as one cluster (20,000 permutations, fixed
seed 27) and report leave-one-goal-out (LOGO) sensitivity.  These are
per-contrast tests; the main paper separately reports Benjamini--Hochberg over
the twelve Bank model contrasts.

\section{Complete Statistical Breakdowns}
\label{app:results}

% Generated by paper/scripts/gen_tables.py; do not edit manually.

\begin{table}[t]
\centering
\small
\setlength{\tabcolsep}{0.8pt}
\begin{tabular*}{\columnwidth}{@{\extracolsep{\fill}}lccccc@{}}
\toprule
Domain & $n$ & flat & \pg{} & W:L & $p$ \\
\midrule
Bank & 125 & 85.6 & 88.8 & 6:2 & .289 \\
Hotel & 195 & 82.1 & 87.2 & 26:16 & .164 \\
Library & 66 & 93.9 & 92.4 & 3:4 & 1.000 \\
DMV & 97 & 92.8 & 94.8 & 5:3 & .727 \\
Healthcare & 124 & 78.2 & 91.9 & 20:3 & $<$.001 \\
Market & 172 & 86.0 & 93.0 & 15:3 & .008 \\
University & 42 & 95.2 & 95.2 & 0:0 & 1.000 \\
Pooled & 821 & 85.7 & 91.1 & 75:31 & $<$.001 \\
\bottomrule
\end{tabular*}
\caption{Plus seven-domain replication. Pass columns are percentages; W:L
counts \pg{}-only versus flat-only successes on the complete paired analysis
set.}
\label{tab:plus-details}
\end{table}

\begin{table}[t]
\centering
\small
\setlength{\tabcolsep}{0.8pt}
\begin{tabular*}{\columnwidth}{@{\extracolsep{\fill}}lccccc@{}}
\toprule
Domain & $n$ & full-cur. & \pg{} & W:L & $p$ \\
\midrule
Bank & 125 & 89.6 & 92.8 & 5:1 & .219 \\
Hotel & 195 & 90.8 & 92.3 & 11:8 & .648 \\
Library & 65 & 96.9 & 95.4 & 0:1 & 1.000 \\
DMV & 96 & 97.9 & 97.9 & 1:1 & 1.000 \\
Healthcare & 124 & 84.7 & 87.9 & 7:3 & .344 \\
Market & 172 & 90.7 & 93.6 & 10:5 & .302 \\
University & 42 & 90.5 & 92.9 & 2:1 & 1.000 \\
Pooled & 819 & 91.0 & 92.9 & 36:20 & .044 \\
\bottomrule
\end{tabular*}
\caption{V4 selective-visibility contrast by domain. Full-cursor keeps the
active frame first and the complete program below it; W:L counts \pg{}-only
versus full-cursor-only successes on the three-arm common set.}
\label{tab:fullcursor-domains}
\end{table}

\begin{table}[t]
\centering
\small
\setlength{\tabcolsep}{1pt}
\begin{tabular*}{\columnwidth}{@{\extracolsep{\fill}}lccccc@{}}
\toprule
Split & $n$ & flat & flat+prot. & full-cur. & \pg{} \\
\midrule
All & 125 & 68.8 & 74.4 & 55.2 & 54.4 \\
Execute & 39 & 46.2 & 23.1 & 12.8 & 10.3 \\
Refusal & 86 & 79.1 & 97.7 & 74.4 & 74.4 \\
\bottomrule
\end{tabular*}
\caption{Exploratory 7B Bank boundary (pass\,\%). Full-cursor and \pg{} are
indistinguishable overall (\pg{}:full-cursor 20:21, $p{=}1$), and both trail
flat. This is boundary evidence, not a pure causal isolation of salience.}
\label{tab:7b-fullcursor}
\end{table}

Tables~\ref{tab:plus-details}--\ref{tab:7b-fullcursor} give the complete new
runtime controls. Plus independently reproduces the strong-model cross-domain
effect. For V4, full-cursor recovers most of the overall and refusal gain while
selective visibility adds a smaller pooled improvement; no individual-domain
\pg{}:full-cursor contrast is significant. The 7B boundary remains
exploratory: retaining the global program does not recover flat performance,
which is consistent with a burden shared by active-frame-guided execution but
does not isolate that component causally.

\begin{table}[t]
\centering
\small
\setlength{\tabcolsep}{0pt}
\begin{tabular*}{\columnwidth}{@{\extracolsep{\fill}}llccccc@{}}
\toprule
Model & Arm & Pass & Exec. & Ref. & Bal. & Seq. \\
\midrule
DS-V4 & flat & 86.4 & 71.8 & 93.0 & 82.4 & 0 \\
             & \pg{} & 92.8 & 76.9 & 100.0 & 88.5 & 0 \\
Q3.7-Plus & flat & 85.6 & 61.5 & 96.5 & 79.0 & 0 \\
             & \pg{} & 88.8 & 69.2 & 97.7 & 83.5 & 0 \\
Q3.6-A3B & flat & 81.6 & 61.5 & 90.7 & 76.1 & 0 \\
             & \pg{} & 84.0 & 71.8 & 89.5 & 80.7 & 1 \\
Q3.6-27B & flat & 86.4 & 64.1 & 96.5 & 80.3 & 1 \\
             & \pg{} & 85.6 & 64.1 & 95.3 & 79.7 & 2 \\
Q2.5-7B & flat & 68.8 & 46.2 & 79.1 & 62.6 & 1 \\
             & \pg{} & 54.4 & 10.3 & 74.4 & 42.3 & 10 \\
4o-mini & flat & 75.2 & 69.2 & 77.9 & 73.6 & 0 \\
             & \pg{} & 48.8 & 23.1 & 60.5 & 41.8 & 19 \\
\bottomrule
\end{tabular*}
\caption{Bank screened subset by class. Bal. is the mean of execute and
refusal accuracy, preventing the 39/86 class ratio from determining the
summary. Seq. counts failures with a correct state/no constraint violation
whose remaining failure is the action-sequence criterion.}
\label{tab:balanced}
\end{table}

\begin{table}[t]
\centering
\small
\setlength{\tabcolsep}{1.5pt}
\begin{tabular*}{\columnwidth}{@{\extracolsep{\fill}}lccc@{}}
\toprule
Pool & \pg{}:flat & Goal $p$ & LOGO max $p$ \\
\midrule
V4 Bank & 9:1 & .191 & .125 \\
V4 seven-domain & 58:19 & .004 & $<$.001 \\
V4 Bank+Hotel & 22:8 & .134 & .119 \\
Plus seven-domain & 75:31 & .004 & $<$.001 \\
27B six-domain & 41:30 & .333 & .519 \\
7B six-domain & 59:237 & $\le$.0001 & $<$.001 \\
GPT-4o-mini Bank & 9:42 & .028 & .001 \\
Plus Bank & 6:2 & .499 & .688 \\
\bottomrule
\end{tabular*}
\caption{Goal-clustered sensitivity. Goal $p$ flips all discordants in a
goal jointly; LOGO is the largest task-level $p$ after dropping one goal.}
\label{tab:goalcluster}
\end{table}

The clustered analysis narrows the claim: Bank-only positive effects are
underpowered at the goal unit, while both cross-domain strong improvements and
weak harm survive.  V4's largest net goal contributions are
Hotel/\texttt{modify\_reservation} (+7),
Healthcare/\texttt{update\_policy} (+7), and Market/\texttt{place\_order}
(+6); for Plus they are Healthcare/\texttt{update\_policy} (+9),
Hotel/\texttt{book\_room} (+7), and Hotel/\texttt{modify\_reservation} (+6).
Dropping any one goal leaves both task-level conclusions significant.

\paragraph{Recorded token and call cost.}
Table~\ref{tab:tokens} aggregates provider-reported usage over every model
call in each DeepSeek-V4-Flash Bank task.  Paging reduces the median system
message from 9,941 to 7,711 characters relative to flat, but does not reduce
the complete prompt after dialog/tool history is included.  It induces more
model calls and substantially more total prompt tokens.  The method is an
accuracy/traceability intervention, not a token-saving result.

\begin{table}[t]
\centering
\small
\setlength{\tabcolsep}{1pt}
\begin{tabular*}{\columnwidth}{@{\extracolsep{\fill}}lccc@{}}
\toprule
Arm & Prompt tokens & Calls & System chars \\
\midrule
offtext & 65,802 [52,503--114,312] & 7 [6--12] & 16,299 \\
flat    & 59,051 [43,607--115,755] & 6 [5--12] & 9,941 \\
\pg{}  & 160,658 [103,114--225,509] & 14 [11--17] & 7,711 \\
\bottomrule
\end{tabular*}
\caption{Bank token/call audit; entries are per-task median [IQR], except
system characters which are per-call.}
\label{tab:tokens}
\end{table}

The mean prompt-token totals are 109,553, 85,867, and 168,602 for offtext,
flat, and \pg{}, respectively.  This accords with the 1.3--2.5$\times$
latency cost disclosed in the main paper and is a material deployment tradeoff.

\section{Mechanism and Ablation Protocols}
\label{app:mechanism}

\paragraph{Representation factorial.}
\emph{off-code} changes only structure/format and retains every official leaf
byte-for-byte; \emph{c-prose} verbalizes the compiled checks, order, gates, and
recipes without code syntax.  A verifier checks 834 leaf correspondences.
The English emission reuses the same leaves and identifiers.  These controls
show directionally stable but incomplete language evidence: they do not
constitute a six-model, seven-domain English \pg{} replication.

\paragraph{Protocol/paging decoupling.}
\emph{flat+protocol} keeps the complete program resident while requiring the
same external bookkeeping actions.  Its contrasts with flat are nonsignificant
at all three sampled tiers; contrasts from flat+protocol to \pg{} retain the
strong positive and weak negative effects.  This isolates visibility from
protocol overhead, but it does not distinguish a stack VM from oracle chunking,
retrieval, a checklist tracker, or an equal-length subtree baseline.  That
missing baseline remains an experimental limitation.

\paragraph{Probe and natural audit.}
The directed probe is machine-graded against saved runtime cursors and tests
whether a model can reconstruct state when explicitly asked.  The natural
audit measures extra domain calls after the target action on unprompted
trajectories.  Its positive region is a reproducible one-sided harm signal,
while the null region drifts across reruns; it is not a calibrated independent
capability scale or deployment threshold.  Attention analyses on two
open-weight endpoints are corroborative and not treated as a causal proof.

\section{Defect and Omission Audit}
\label{app:defects}

The data-selection rule was fixed before arm comparisons: omit execute tasks
whose goal had zero success in the upstream trajectory set used during setup;
retain refusal tasks because they do not require the target transition.  The
rule is reproducible, but a zero-pass observation is not by itself proof of an
unreachable simulator state.  We therefore ran a deterministic current-source
oracle replay and now distinguish confirmed defects from empirical omissions.

\begin{table}[t]
\centering
\small
\setlength{\tabcolsep}{1.2pt}
\begin{tabular}{@{}p{.13\columnwidth}p{.41\columnwidth}p{.12\columnwidth}p{.28\columnwidth}@{}}
\toprule
Domain & Goal (upstream) & Replay & Classification \\
\midrule
Bank & \shortstack[l]{\texttt{cancel\_credit\_card}\\(0/300)} & no & confirmed list/dict iteration defect \\
Bank & \shortstack[l]{\texttt{pay\_bill\_with\_}\\\texttt{credit\_card} (0/100)} & no & true return without required mutation \\
Bank & \texttt{open\_account} (8/50) & yes & empirical/version-sensitive omission \\
Library & \texttt{add\_book} (0/48) & yes & empirical omission; no confirmed defect \\
DMV & \shortstack[l]{\texttt{update\_test\_status}\\(0/49)} & yes & empirical omission; no confirmed defect \\
\bottomrule
\end{tabular}
\caption{Audit of all screened goals. Parentheses give upstream
pass/execute counts; replay tests the target transition after prerequisites.}
\label{tab:defects}
\end{table}

For \texttt{cancel\_credit\_card}, the task database stores credit cards as a
list of records, while the action iterates the container as if keys were card
numbers; comparison with the requested string therefore fails.  For
\texttt{pay\_bill\_with\_credit\_card}, the same representation mismatch lets
the method return true without increasing the matching card's balance, so the
oracle final state cannot match.  In contrast, current-source replay creates
the requested account, adds the requested book, and updates the test status
with a driver's license.  We do not claim those three are impossible.

The main Bank subset omits nine execute instances before any arm is run.
Treating all nine as failures over the original denominator yields
offtext/flat/\pg{} = 65.7/80.6/86.6, preserving the ordering; since the same
instances are absent from every arm, the paired discordant counts among
observed instances are unchanged.  Library and DMV each omitted one execute
instance before collection; their domain cells therefore apply to the disclosed
screened set, and no imputed significance claim is made.  This asymmetry is a
limitation rather than evidence for a third, fourth, or fifth defect.

\section{Qualitative Industrial Portability Case}
\label{app:industrial}

With authorization and de-identification, we include a supply-chain quotation
workflow instantiated on the same PG representation and stack runtime outside
SOPBench.  The current program contains 13 functions and 87 labeled steps,
references four domain adapters plus two generic runtime tools, and required no
customer-specific runtime modification.  The PG was manually engineered from
business rules and spreadsheet/tool contracts; it was \emph{not} generated by
the benchmark graph compiler or an unevaluated natural-language front end.

An employee-run task on this exact simplified snapshot produced a 24-event
runtime trace from \texttt{start} through \texttt{complete}.  The typed profile,
output plan, and rendered workbook validations all passed; the final workbook,
render report, and 560-entry calculation trace were present in the task's
generated-result directory (not its upload directory), with no runtime error or
unresolved item.  The submitted event projection retains only action,
function, and step fields.  Raw client workbooks, commercial adapters, paths,
messages, and hidden reasoning are excluded; an English synthetic workbook with
91 formulas and seven passing consistency checks exposes the dataflow without
releasing business values.

This case establishes qualitative representation/runtime portability and
quantifies the domain engineering surface.  It is not a PG-versus-flat
comparison, does not estimate accuracy, latency, labor, or cost gains, and does
not validate the capability-gate threshold.

\section{Artifact Contents and Verification}
\label{app:artifact}

The anonymous archive contains the compiler, runtime snapshot and PG extension,
adapter, aggregate/per-task result JSONs used by the paper, generated programs,
analysis scripts, configuration templates without credentials, and a SHA-256
manifest.  It also contains the English, de-identified industrial case described
above: the current PG, tool-interface contracts,
payload-free runtime events, validation summary, and synthetic review workbook.
Large raw provider traces and conversation states are omitted for
privacy and size; their anonymous aggregates are included. SOPBench is a
public dependency pinned to commit prefix \texttt{b30bb29c3438} (the full SHA
is in \texttt{COMMIT\_PIN.txt}); the review archive includes
only the minimal compatibility metadata needed by the scripts rather than the
full public benchmark checkout.  The archive audit rejects credentials,
absolute author paths, identity strings, hidden metadata, and non-whitelisted
result roots before building the ZIP.

\end{document}